\documentclass[sigconf]{acmart}

\AtBeginDocument{%
  }

\copyrightyear{2023}
\acmYear{2023}
\setcopyright{rightsretained}
\acmConference[MM '23]{Proceedings of the 31st ACM International Conference on Multimedia}{October 29-November 3, 2023}{Ottawa, ON, Canada}
\acmBooktitle{Proceedings of the 31st ACM International Conference on Multimedia (MM '23), October 29-November 3, 2023, Ottawa, ON, Canada}
\acmDOI{10.1145/3581783.3613462}
\acmISBN{979-8-4007-0108-5/23/10}

\begin{document}

\title{HypLL: The Hyperbolic Learning Library}

\author{Max van Spengler}
\affiliation{%
  \institution{University of Amsterdam}
  \country{the Netherlands}
  \texttt{m.w.f.vanspengler@uva.nl}
}

\author{Philipp Wirth}
\affiliation{%
  \institution{Lightly}
  \country{Switzerland}
  \\
  \texttt{philipp@lightly.ai}
}

\author{Pascal Mettes}
\affiliation{%
  \institution{University of Amsterdam}
  \country{the Netherlands}
  \texttt{p.s.m.mettes@uva.nl}
}

\renewcommand{\shortauthors}{Max van Spengler, Philipp Wirth, \& Pascal Mettes}

\begin{abstract}
Deep learning in hyperbolic space is quickly gaining traction in the fields of machine learning, multimedia, and computer vision. Deep networks commonly operate in Euclidean space, implicitly assuming that data lies on regular grids. Recent advances have shown that hyperbolic geometry provides a viable alternative foundation for deep learning, especially when data is hierarchical in nature and when working with few embedding dimensions. Currently however, no accessible open-source library exists to build hyperbolic network modules akin to well-known deep learning libraries. We present HypLL, the Hyperbolic Learning Library to bring the progress on hyperbolic deep learning together. HypLL is built on top of PyTorch, with an emphasis in its design for ease-of-use, in order to attract a broad audience towards this new and open-ended research direction. The code is available at: \url{https://github.com/maxvanspengler/hyperbolic\_learning\_library}.
\end{abstract}



\maketitle

\section{Introduction}
Deep learning plays a central role in Artificial Intelligence on any data type and modality. Advances in deep learning are distilled in well-known and broadly used software libraries, such as PyTorch~\cite{paszke2019pytorch}, Tensorflow~\cite{tensorflow2015-whitepaper}, and MxNet~\cite{chen2015mxnet} amongst others. Contemporary deep learning libraries are implicitly or explicitly designed for Euclidean operators, with optimized matrix and vector operators and corresponding automatic differentiation. The underlying assumption is that data is best represented on regular grids. In practice however, this assumption might not be sufficient, desirable, or even workable \cite{bronstein2017geometric}.

One common limitation of conventional differentiable deep learning algorithms is dealing with hierarchical data. Hierarchies are ubiquitous, both in our data and in the semantics we seek to derive from our data. Hierarchies also form a foundation across all sciences for distilling and abstracting knowledge \cite{noy1997state}. In hierarchies, the number of nodes generally grows exponentially with depth. On the other hand, the volume of a ball in Euclidean space grows polynomially with diameter, leading to distortion when embedding hierarchies \cite{bachmann2020constant}. For embedding hierarchies, foundational work in hyperbolic learning has shown that hyperbolic space is superior, leading to minimal distortion while requiring only few embedding dimensions \cite{ganea2018hyperbolic2,nickel2017poincare}. 

The advances in hyperbolic embeddings of hierarchies and subsequent theory on hyperbolic network layers \cite{ganea2018hyperbolic,shimizu2021hyperbolic} have led to rapid developments towards hyperbolic deep learning across many modalities. Hyperbolic learning algorithms have been proposed for graphs \cite{liu2019hyperbolic}, text \cite{tifrea2019poincar}, computer vision \cite{atigh2022hyperbolic,ermolov2022hyperbolic,khrulkov2020hyperbolic}, recommender systems \cite{wang2021fully} and more. The growing body of literature has already been captured in multiple surveys \cite{mettes2023hyperbolic,peng2021hyperbolic,yang2022hyperbolic}, we kindly refer to these works for detailed descriptions of ongoing literature. Across these works, hyperbolic learning has shown great potential, with improvements when data is hierarchical or scarce, with more robustness to out-of-distribution and adversarial samples, and better performance with few embedding dimensions.

An open challenge in hyperbolic learning is the lack of shared open source implementation and development. With current deep learning libraries centered around Euclidean operators, building upon existing frameworks is not directly feasible for hyperbolic learning. Multiple leading repositories have already been made for optimization on hyperbolic and other non-Euclidean spaces \cite{2106.08777,geoopt2020kochurov,JMLR:v21:19-027}.
Such approaches do however not keep track of underlying manifolds or have implementations of common network layers. To fill the gap in current literature, we present HypLL: The Hyperbolic Learning Library. Our repository is built upon PyTorch and contains implementations of well-known network layers. We take two core principles to heart: (i) our functionality should closely resemble PyTorch and (ii) it should be easy to use and debug, even in the presence of different manifolds. The library is targeted both for researchers in hyperbolic deep learning, as well as AI practitioners broadly, without needing all the mathematical knowledge before getting started in this research direction.

\section{Hyperbolic Learning Library design}

Central in the design of HypLL is making the step towards hyperbolic learning easy for PyTorch users and to keep the library easy to use, even when dealing with many manifolds.
In pure PyTorch (and any other library), it can become difficult and tedious during computations to keep track of the manifold and other metadata underlying the data within tensors. Especially in deep learning, when the number of tensors in the computational graph increases rapidly, this problem becomes challenging. As a result, mistakes happen frequently and tend to be difficult to spot and correct. We have built our design around keeping track of manifolds, to make the network design transparent and easy to debug. In contrast, while Hyperlib also provides hyperbolic learning functionalities \cite{hyperlib}, it is only available for Tensorflow, does not keep track of manifolds, and does not contain important layers such as convolutions and batch normalization.

\begin{figure}
    \centering
    \includegraphics[width=0.45\textwidth]{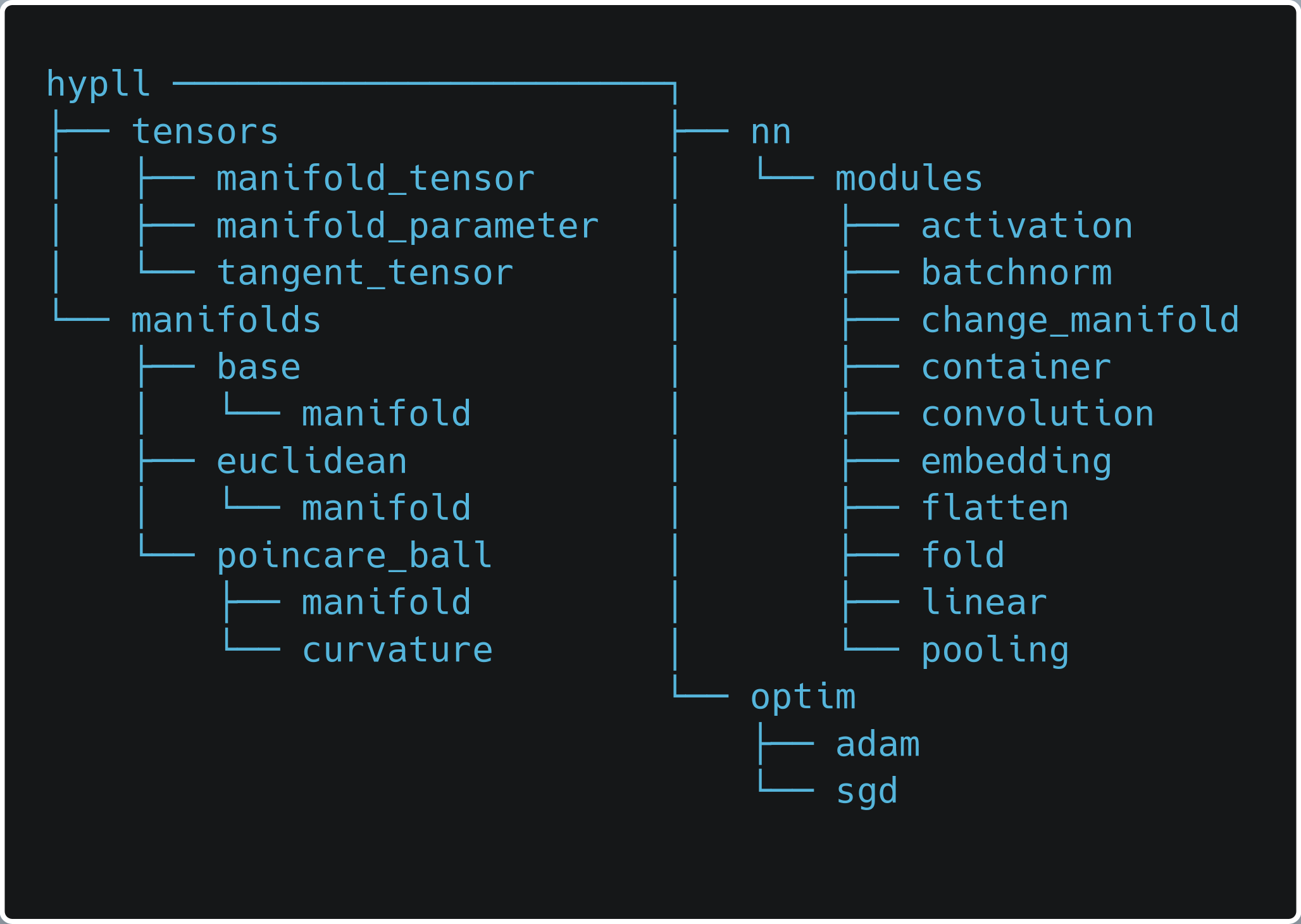}
    \caption{The structure of HypLL, centered around the tensors, manifolds, nn, and optim modules.}
    \label{fig:lib_design}
\end{figure}

In this Section, we will highlight the design of the core modules in HypLL. The overall structure of the library is shown in Figure \ref{fig:lib_design}. The library is centered around four modules: (i) the tensors module, (ii) the manifolds module, (iii) the nn module, and (iv) the optim module. The modules are discussed sequentially below.

\subsection{The tensors module}
The tensors module forms the foundation and contains three important components: 1) the manifold tensor, 2) the manifold parameter, and 3) the tangent tensor. The first and third of these take over a part of the role of the original tensor class from the PyTorch library. The manifold parameter takes over the role of the PyTorch parameter class. As such, these classes form the basic objects for storing data throughout any computations with the other modules.

\textbf{The manifold tensor} is a class with three important properties: 1) a PyTorch tensor, 2) a manifold $\mathcal{M}$ from the manifold module which will be discussed in Subsection \ref{subect:manifolds} and 3) a manifold dimension $d \in \mathbb{Z}$. The manifold indicates where the data --- stored in the tensor object --- lives. The manifold dimension indicates the dimension that stores the points on the manifold. For example, if we have a 2-dimensional manifold tensor with a Poincar\'e ball manifold and manifold dimension 1, then each row in our tensor contains a point on the Poincar\'e ball. By storing this additonial data on the manifold tensor, we can later ensure that any operation applied to the manifold tensor is indeed allowed. For example, if an operation assumes data to be Euclidean, while the data is actually hyperbolic, we can easily point out the mistake.

Not all data lives on a manifold. For example, a tensor with label indices does not have an underlying manifold. In such cases we revert to the tensor class from PyTorch. Hence in HypLL, this class bears a slightly different interpretation; a tensor containing values which do not form vectors or points on a manifold.

\textbf{The manifold parameter} is simply a manifold tensor which subclasses the parameter class from PyTorch. This allows creating layers with points on a manifold as its parameters, which will prove important when discussing the nn module.

\textbf{The tangent tensor} is similar to the manifold tensor in that it stores metadata for a collection of data stored in its tensor attribute. However, here the data consists of vectors living in the tangent space of the manifold $\mathcal{M}$ that is within the manifold attribute of the tangent tensor. A tangent space, written as $\mathcal{T}_x \mathcal{M}$, is defined by a manifold $\mathcal{M}$ and a point $x \in \mathcal{M}$. 
When working in hyperbolic space, it is convenient to have tangent vectors from various tangent spaces $\{\mathcal{T}_{x_i} \mathcal{M}\}_i$, stored in the same tangent tensor. Therefore, the tangent tensor also contains a manifold tensor which has the same manifold and for which the tensor attribute is broadcastable with the tensor of tangent vectors. Allowing broadcastable tensors instead of tensors of the same shape makes these tangent tensors more flexible while reducing memory requirements. If this manifold tensor is set to \texttt{None}, every tangent vector is located at the origin of the manifold. Lastly, the tangent tensor contains a manifold dimension, which is again an integer indicating what dimension contains the vectors.

To summarize, the tangent tensor contains a tensor attribute containing tangent vectors, a manifold attribute indicating the manifold to which the vectors are tangent, a manifold tensor attribute containing the points on the manifold where the tangent spaces are located and is broadcastable with the tangent vectors; and a manifold dimension indicating the dimension of the vectors.

\subsection{The manifolds module}\label{subect:manifolds}
The manifolds module contains the different manifolds that the library currently supports. These classes contain all of the usual operations that are defined on these manifolds and which are required for the operations defined in the nn module. Each different manifold subclasses the base manifold class, which is a metaclass containing the methods that each manifold should have.
In the current implementation, we have focused on the Euclidean manifold and the Poincar\'e ball, the most commonly used model of hyperbolic space. The library is designed to be able to incorporate any other manifold as well in future updates, such as the hyperboloid and Klein models. 
The inclusion of the Euclidean manifold within this module is required for optimally providing flexible manifold-agnostic layers in the nn module.

\begin{figure*}
    \centering
    \includegraphics[width=0.9\textwidth]{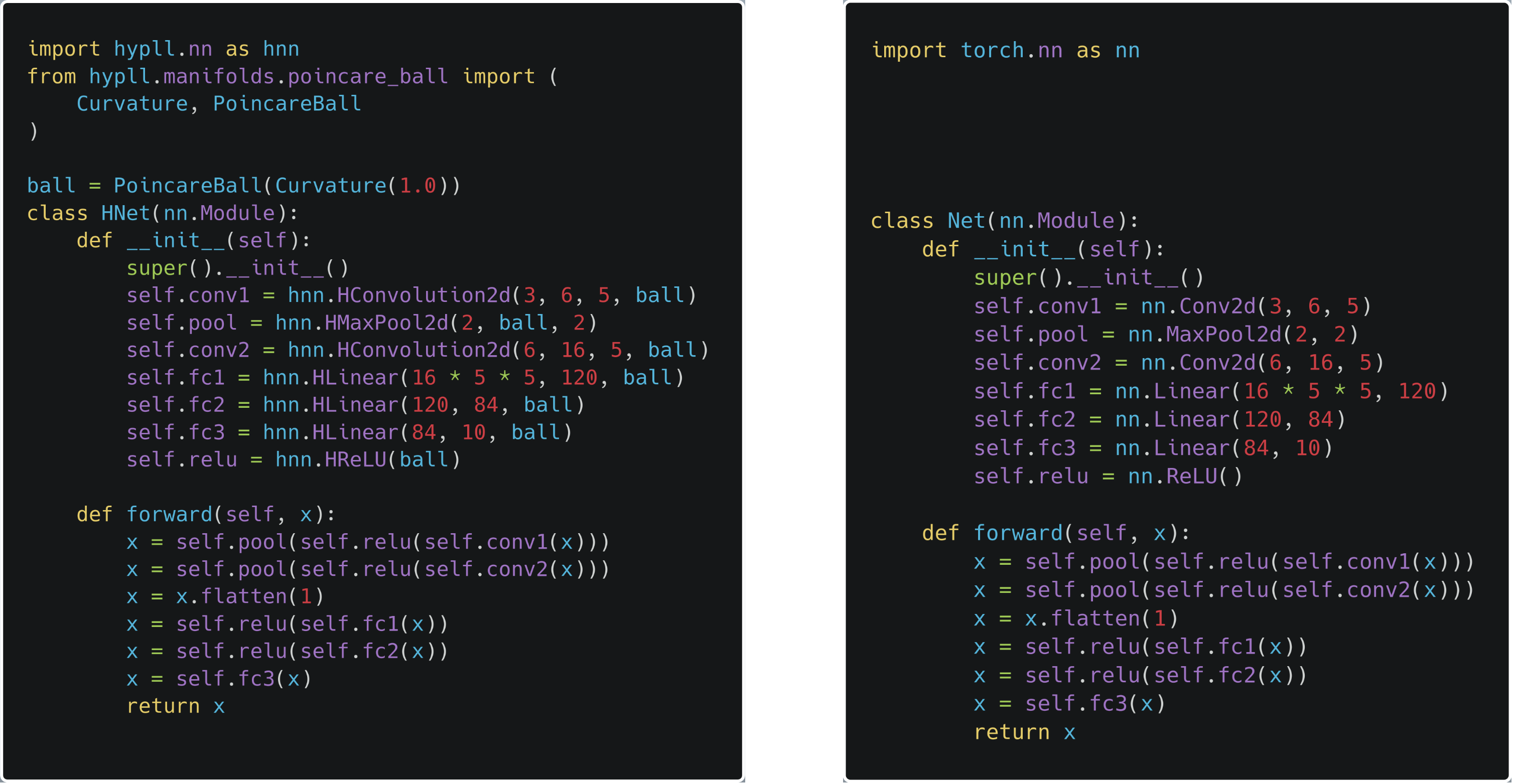}
    \caption{Comparison of the implementation of a small convnet in HypLL (left) versus in PyTorch (right).}
    \label{fig:model_comparison}
\end{figure*}

Each manifold submodule contains the mathematical functions that define its operations, such as exponential maps, logarithmic maps, the Fr\'echet mean, vector addition and more. These operations apply checks to see if their inputs live on the correct manifold and, when there are multiple inputs, if the manifold dimensions of the inputs align properly. This is the largest contributor to our second design principle, as these operations significantly reduce the difficulty of debugging by explicitly disallowing mathematically nonsensical operations. Without such checks, operations can easily lead to silent failures and tricky bugs. For a complete overview of the different operations defined on these manifolds, with implementations based on \cite{ungar2008gyrovector,ganea2018hyperbolic, shimizu2021hyperbolic,geoopt2020kochurov}, we refer to the source code.

Another important component of this module is the curvature of the Poincar\'e ball manifold. The curvature is a module containing a single parameter, which is used to compute the absolute value of the curvature and can be made learnable. The reason to use the absolute value of the curvature instead of the true negative value is to avoid having to add a minus sign throughout, which increases ease-of-use. This does not lead to down-stream issues as we only support non-positive curvature manifolds. Such a curvature object is supplied as input to the Poincar\'e ball class during initialization to define the curvature of the manifold.

\subsection{The nn module}
The nn module is where all of the neural network methodology is implemented. It is structured similarly to the nn module from PyTorch and contains the currently available learning tools for Euclidean space and the Poincar\'e ball model. Similar to the classes in the PyTorch nn module, each of the classes in our nn module subclasses the Module class from PyTorch, which ensures that they will be properly registered for optimization. This module will be expanded whenever new methodology becomes available in literature. An overview of the available layers is shown in Figure~\ref{fig:lib_design}. The implementations are based on \cite{ganea2018hyperbolic,shimizu2021hyperbolic,lou2020differentiating,nickel2017poincare,vanspengler2023poincare}.

Each of the layers in the nn module is designed to be manifold-agnostic. In practice, each layer is supplied with a manifold object and it uses the operations defined on this manifold object to define its forward pass. So, when the supplied manifold is Euclidean space, the layers are equivalent to their PyTorch counterparts. Due to the usage of these manifold operations, all of the compatibility checks on the inputs are automatically built-in, which increases ease-of-use.
Following our first design principle, this manifold argument that is supplied to each layer is the only difference between the signature of the original PyTorch layers and our layers. As a result, the only difference between building a neural network with HypLL compared to with PyTorch is having to define a manifold and supplying this manifold to the layers of the network.

\subsection{The optim module}
The optim module implements the Riemannian SGD and Riemannian Adam optimizers as introduced in \cite{becigneul2018riemannian}, based on the implementations from \cite{geoopt2020kochurov}. These implementations work both with manifold parameters and PyTorch's parameters. When optimizing manifold parameters, the optimizers use the manifold of the manifold parameter for each of the operations that is performed during optimization. As a result, the checks are again built-in automatically through the manifold objects. When training with manifold parameters on the Euclidean manifold or with PyTorch parameters, the optimizers are equivalent to their PyTorch counterparts. Following our first design principle, initialization of these optimizers is identical to the optimization of the PyTorch optimizers. Moreover, these optimizers inherit from the base PyTorch optimizer, which makes them compatible with learning rate schedulers and other such tools.

\section{Example usage}\label{sec:example_usage}
To showcase how easy it becomes to define and train a neural network with HypLL, we will describe the similarities and differences with the usage of PyTorch.
The major differences that come with using our library are 1) defining a manifold on which our data will live and on which our model will act; and 2) moving our input data onto this manifold as a pre-processing step. For this example we will use the CIFAR-10 tutorial from the PyTorch documentation\footnote{https://pytorch.org/tutorials/beginner/blitz/cifar10\_tutorial.html (07-06-2023)}, which we have also adapted as a tutorial for our library. We will show several steps involved in training this small convolutional network and compare it to the PyTorch implementation.

\textbf{Creating a network.}
We start by defining the manifold and then using this manifold to define a small convolutional network. We will use a Poincar\'e ball with curvature -1 for this example. The implementations of this network in HypLL and in PyTorch are shown side-by-side in Figure \ref{fig:model_comparison}. The only true difference is that we define a manifold and supply it to the model layers in the HypLL code. Adding hyperbolic geometry to a network is as simple as that with this library.

\begin{figure}
    \centering
    \includegraphics[width=0.35\textwidth]{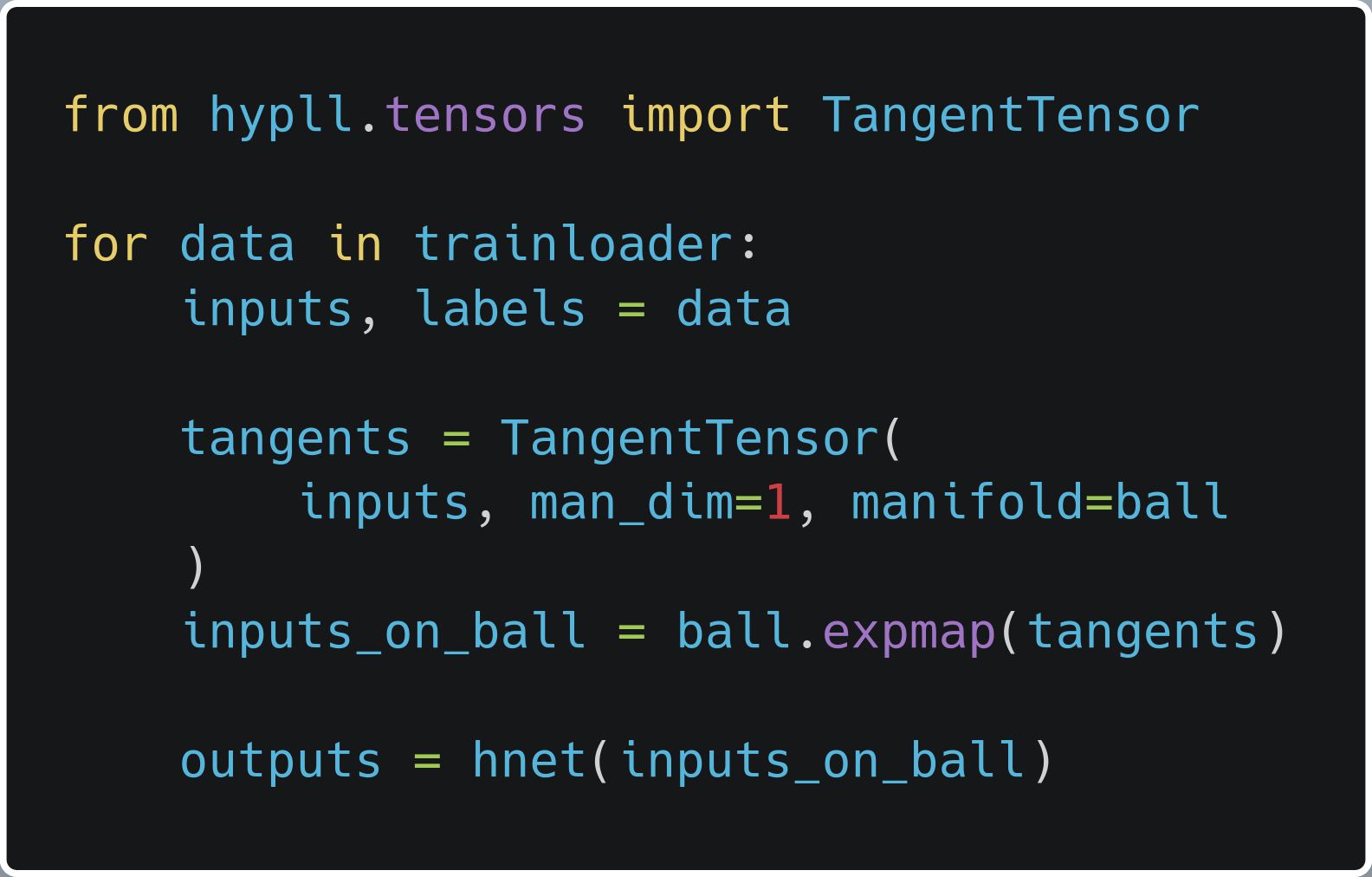}
    \caption{Feeding data to a model written with HypLL.}
    \label{fig:training}
\end{figure}

\textbf{Feeding data to our model.}
Second, we show part of the training loop, where we only show the part in which our implementation differs from PyTorch for brevity. Mapping Euclidean vectors to hyperbolic space is usually performed by assuming the vectors to be tangent vectors at the origin and then mapping these to the hyperbolic space using the exponential map. We will use this approach here as well. The example is shown in Figure \ref{fig:training}.

So, when using HypLL, a little bit of logic has to be added to move the inputs to the manifold on which the network acts. Namely, we first wrap the inputs in a tangent tensor with no manifold tensor argument and then map it using the exponential map of the Poincar\'e ball. This operation is left to the user so they have full control over how to move their inputs to hyperbolic space. Aside from that, nothing else changes, making hyperbolic deep learning a tool that can be used by a broad audience. 

\section{Conclusions and outlook}
This paper presents the Hyperbolic Learning Library, enabling researchers and practitioners to perform deep learning without hassle in hyperbolic space, a new and open-ended research direction.
HypLL is designed to make the step from PyTorch minimal and to keep debugging easy by tracking manifolds. The library is a continual effort, where the latest advances in the field are continually integrated and forms a central point to work on challenges in the field, such as increasing stability in optimization and performing learning at large scale. The main structure follows the ideology of PyTorch \cite{paszke2019pytorch} with corresponding modules. In the future, we strive to build a geometric framework on top of the library for graph-based data, in the spirit of PyG, PyTorch geometric \cite{fey2019fast}.

\section{Acknowledgements}
Max van Spengler acknowledges the University of Amsterdam Data Science Centre for financial support.

\bibliographystyle{ACM-Reference-Format}
\bibliography{acmart}

\end{document}